\pdfoutput=1 
\documentclass{article}

\PassOptionsToPackage{numbers, compress}{natbib}

\usepackage[preprint]{neurips_2019}

\usepackage[utf8]{inputenc} 
\usepackage[T1]{fontenc}    
\usepackage{hyperref}       
\usepackage{url}            
\usepackage{booktabs}       
\usepackage{amsfonts}       
\usepackage{nicefrac}       
\usepackage{microtype}      
\usepackage{color}      

\usepackage{microtype}
\usepackage{graphicx}
\usepackage{subfigure}
\usepackage{booktabs} 
\usepackage{amsthm}
\usepackage{amsmath}
\usepackage{float}
\usepackage{wrapfig}
\usepackage{makecell}
\usepackage{caption}
\usepackage{bm}
\usepackage{microtype}
\usepackage{mathtools}
\usepackage{multicol}
\usepackage{multirow}
\usepackage{amsmath}
\usepackage{amsfonts}
\usepackage{makecell}
\usepackage{comment}
\usepackage{dblfloatfix}
\usepackage{enumitem}
\usepackage[linesnumbered, ruled]{algorithm2e}
\usepackage[symbol]{footmisc}

\usepackage{titlesec}

\titlespacing{\section}{0pt}{0.5\parskip}{0.5\parskip}
\titlespacing{\subsection}{0pt}{0.5\parskip}{0.5\parskip}
\titlespacing{\subsubsection}{0pt}{0.5\parskip}{0.5\parskip}
\titlespacing{\paragraph}{0pt}{0.5\parskip}{0.5\parskip}

\SetKwRepeat{Do}{do}{while}%
\SetKwInOut{Parameter}{Parameter}

\DeclareMathOperator*{\argmax}{arg\,max}

\usepackage{chngcntr}
\usepackage{caption} 
\title{Doubly Sparse: Sparse Mixture of Sparse Experts \\ for Efficient Softmax Inference}

\author{%
  Shun Liao \\
  University of Toronto\\
  \And
  Ting Chen \\
  Google \\
  \And
  Tian Lin \\
  Google \\ 
  \And
  Denny Zhou \\
  Google \\
  \And
  Chong Wang \\
  ByteDance
}

\begin{document}

\maketitle
\begin{abstract}
Computations for the softmax function are significantly expensive when the number of output classes is large. In this paper, we present a novel softmax \textit{inference} speedup method, Doubly Sparse Softmax (DS-Softmax), that leverages sparse mixture of sparse experts to efficiently retrieve top-k classes. Different from most existing methods that require and approximate a fixed softmax, our method is learning-based and can adapt softmax weights for a better inference speedup. In particular, our method learns a two-level hierarchy which divides entire output class space into several partially overlapping experts. Each expert is sparse and only contains a subset of output classes. To find top-k classes, a sparse mixture enables us to find the most probable expert quickly, and the sparse expert enables us to search within a small-scale softmax. We empirically conduct evaluation on several real-world tasks, including neural machine translation, language modeling and image classification, and demonstrate that significant computation reductions can be achieved at no performance loss. 
\end{abstract}
\section{Introduction}
Deep learning models have demonstrated impressive performance in many classification problems~\citep{lecun2015deep}. In many models, the softmax function is commonly used to produce categorical distributions over the output space. Due to its linear complexity, the computation for the softmax layer becomes a bottleneck with large output dimensions, such as language modeling \citep{bengio2003neural}, neural machine translation \citep{bahdanau2014neural} and face recognition \citep{sun2014deep}. In language modelling task, softmax contributes to more than 95\% of computation for the small model \citep{zaremba2014recurrent, merity2016pointer}. This becomes a significant bottleneck with limited computational resource, such as deploying the model to mobile devices \citep{howard2017mobilenets}.

Many methods have been proposed to reduce softmax complexity. The softmax computation bottleneck is present in both training and inference phases, but there are different objectives. For training, the goal is to estimate the categorical distribution and approximate the normalization term as quick as possible \citep{gutmann2012noise}. Unlike training, the goal in inference is to search for top-k classes accurately and efficiently. Most existing methods formulate this as an approximated maximum inner product search problem: given an already learned/fixed softmax, how to search the top-k classes without linear complexity, noted as post-approximation methods here \citep{shrivastava2014asymmetric, shim2017svd, zhang2018navigating, chen2018learning}. However, the standard learned and fixed softmax may not be structured in a (hierarchical) way such that locating top-k can be easily achieved, which leads to sub-optimal trade-offs between efficiency and accuracy \citep{chen2018learning}. 

In this work, we propose a novel Doubly Sparse Softmax (DS-Softmax), which can make the top-k searching efficiently the inference phase. The DS-Softmax is a learning-based method and adapt softmax to be hierarchically structured during training, which can achieve a better trade-off. The model learns a two-level overlapping hierarchy using {\it sparse} mixture of {\it sparse} experts structure during its training. Each expert is sparse and only contain a small subset of entire output class space, while each class is permitted to belong to more than one expert. Given an input vector and a set of experts, the DS-Softmax first selects the top expert that is most related to the input (in contrast to a dense mixture of experts). Then, the single selected expert can return the categorical distribution on a small subset of classes. Therefore, the reduction is achieved as the model does not need to consider the whole vocabulary. Due to our design, DS-Softmax is also orthogonal with post-approximation methods, so they can be applied to DS-Softmax by treating each expert as another softmax to approximate. Furthermore, training of mixture of experts model can introduce a memory bottleneck. We propose one novel mitosis training scheme to relieve this bottleneck, by progressively cloning from small number of experts to more. 

We conduct experiments in one synthetic dataset and three different real tasks, including language modeling, neural machine translation, and image classification. We demonstrate our method can reduce softmax computation dramatically without loss of prediction performance. For example, we achieved more than 23x speedup in language modeling and 15x speedup in translation without loss of performance. By combing SVD-Softmax, DS-Softmax achieve 32x speedup with similar performance in language modelling. Qualitatively, we demonstrate the learned two-level overlapping hierarchy is semantically meaningful on natural language modeling tasks. By applying mitosis training, the memory requirement for training a 64 experts model is reduced to 4 fold rather than 64.

\textbf{Contributions}. We propose a novel and learning-based method to speedup softmax inference for large discrete output space. To our knowledge, the proposed method the novel method that adapts softmax weights for top-k retrieval instead of simply approximating the softmax distribution. We further propose mitosis training to reduce the memory consumption during the training of our model. Through comprehensive experiments, we show the proposed method provides significant inference speedup for softmax without performance loss.


\section{Related Work}

Reducing the computation cost of softmax has been a long standing problem and widely studied before \citep{gutmann2012noise,chen2015strategies,grave2016efficient,shim2017svd,zhang2018navigating,chen2018learning}. There are mainly two goals: training speedup and inference speedup. In our work, we focus on inference, where we aim to find the top-k classes efficiently and accurately.

\textbf{Post-approximation based}. Most existing works for reducing the softmax inference complexity are based on \textit{post-approximation of a fixed softmax} that has been trained in a standard procedure. Locality Sensitive Hashing (LSH) has been demonstrated as a powerful technique under this category \citep{shrivastava2014asymmetric, maddison2014sampling, mussmann2017fast, spring2017new}. Small word graph is another powerful technique for this problem \citep{zhang2018navigating}. Recent work proposes one learning-based clustering for trained embedding which overcomes the non-differential problem \citep{chen2018learning}. In addition, decomposition-based method, SVD-softmax \citep{shim2017svd}, can speedup the searching through one smaller preview matrix. However, as an approximation to a fixed softmax, the main drawback is that it always suffers high cost when high precision is required \citep{chen2018learning}, suggesting a worse trade-off between efficiency and accuracy. In contrast, the proposed DS-softmax is able to adapt the softmax and learn a hierarchical structure to find top-k classes adaptively. Furthermore, it is possible that those methods can also be applied upon our method, where each expert can be viewed as a single softmax.

\textbf{Hierarchy based}. Another family of related methods try to \textit{incorporate hierarchical structures into softmax}. The most related ones under this category are D-softmax \citep{chen2015strategies} and adaptive-softmax \citep{grave2016efficient}. These two methods can speedup both training and inference while other methods  \citep{morin2005hierarchical, mnih2009scalable} cannot speedup inference. The construction of hierarchy is through unbalanced word/class distribution due to Zipf's law. There are two major issues. Firstly, their hierarchy is pre-defined by heuristics that could be sub-optimal. Secondly, the skewness of class distribution in some tasks, e.g. image classification, may not be as significant as in language modeling. DS-softmax overcomes those limitations by automatically learn the two-level overlapping hierarchy.

\textbf{Mixture of Experts}. Our method is inspired by sparsely-gated mixture-of-experts (MoE) \citep{shazeer2017outrageously}. MoE achieves better performance in language modeling and translation with large but sparsely activated experts. However, MoE cannot speedup the softmax inference by definition because each expert covers the whole output classes. Our work on softmax inference speedup can also be considered as a part of recent efforts to make a neural network more compact \citep{han2015deep,chen2018learningk} and efficient \citep{howard2017mobilenets,chen2019adaptive}, through which we could make modern neural networks faster and more applicable. Also, training MoE suffers the memory bottleneck, this problem is relieved through our novel mitosis training scheme. 
\section{Method}

\begin{figure}[!t]
    \centering\includegraphics[width=0.9\textwidth]{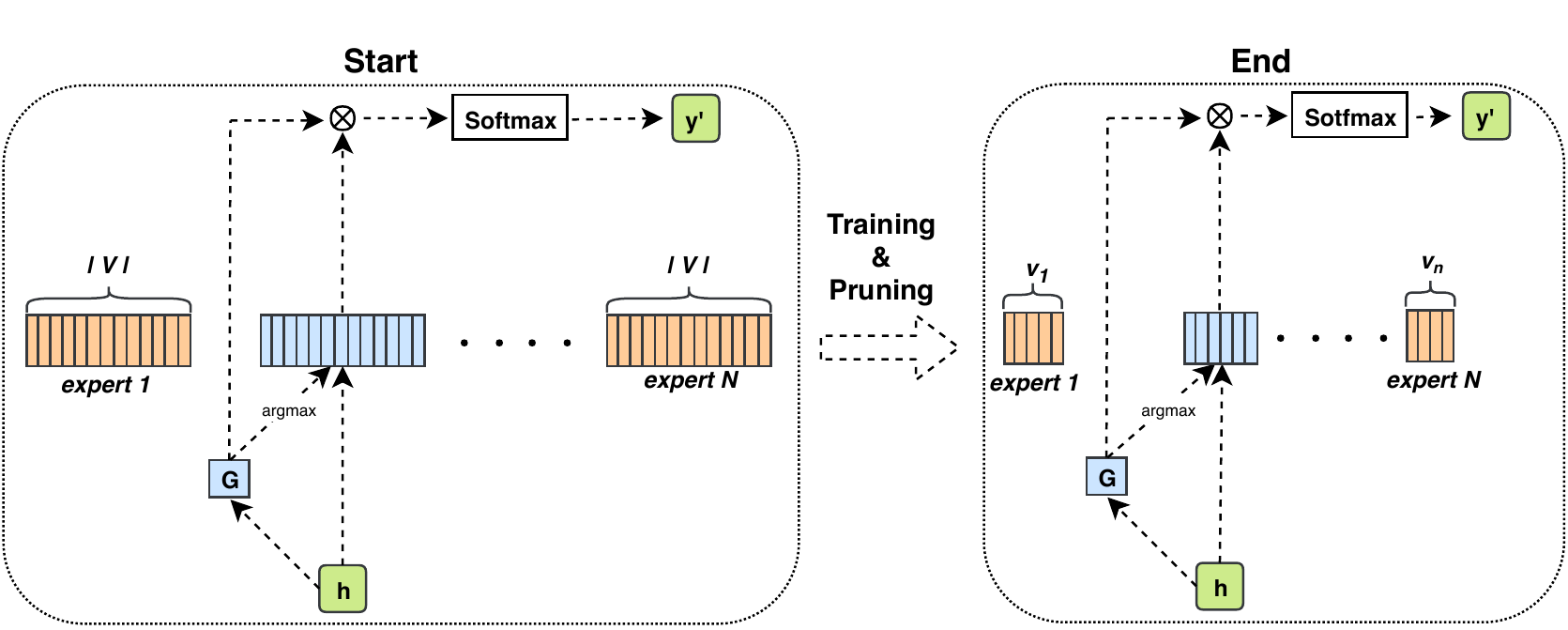}\caption{Overview of Doubly Sparse Softmax (DS-Softmax). Each expert is initialized with the full output space and only the expert with the highest gating value is selected feed-forward. During training, each expert is pruned iteratively so that it only contains a subset of classes, $|v_n|$, in the final model. Therefore, a faster inference can be achieved by only search top-k inside such a subset.}\label{fig:overview_fig}
\end{figure}

In this section, we introduce the softmax inference problem, as well as the proposed method.

\subsection{Softmax Inference Problem}

Given a context vector $h \in \mathbb{R}^d$, a softmax layer is used in order to compute a categorical distribution over a set of classes. In particular, it is defined as $P(class=c|h) = \exp(W_ch)/Z$ where $Z=\sum_i\exp(W_ih)$ is the normalization term and $W\in \mathbb{R}^{N\times d}$ is the softmax embedding parameter. For inference, our goal is not to compute the full exact distribution, but rather to find the top-k classes, i.e. $\{c|P(class=c|h)\ge p_k\}$ where $p_k$ is the $k$-th largest value of $P(class|h)$. The most conventional method to do so is to compute the whole $P(class|h)$ vector and find the top-k, which has $\mathcal{O}(N)$ complexity\footnote{Top-k selection requires an extra $\mathcal{O}(k \log k)$ by Quickselsort.}. Facing a large output space (i.e. large $N$), the softmax layer becomes a bottleneck, and our goal is to find the top-k both accurately and efficiently.

\subsection{Motivation}

Many natural discrete objects/classes, such as natural language, exhibit some hierarchical structure where objects are organized in a tree-like fashion. A hierarchical structure can enable retrieving objects in a much faster way since we do not need to consider the whole set. A two-level hierarchy is studied for language modeling, where each word belongs to a unique cluster while the hierarchy is constructed with different approaches (A ``cluster'' here refers to a cluster of words) \citep{goodman2001classes, chen2015strategies, grave2016efficient}.  However, the construction of the hierarchy is very challenging and usually based on heuristics. Also, it can be very limiting to construct a hierarchy that contains mutual exclusive clusters. This is because, such as in language modeling, it is often difficult to exactly assign a word to a single cluster. For example, if we want to predict the next word of ``I want to eat $\rule{0.5cm}{0.15mm}$'' and one possible correct answer is ``cookie'', we can quickly notice that possible answer belongs to something eatable. If we only search for the right answer inside words with the eatable property, we can dramatically increase the efficiency. Even though words like ``cookie'' are one of the correct answers, it might also appear under some non-edible context such as ``a piece of data'' in computer science literature. Thus, a two-level overlapping hierarchy can naturally accommodate word homonyms like this by allowing each word to belong to more than one cluster. We believe that this observation is valid in other applications besides language modeling.

\subsection{The Doubly Sparse Softmax}
Inspired by such hierarchical structures, we propose our method, Doubly Sparse Softmax (DS-Softmax), to automatically capture and leverage that for softmax inference speedup. The proposed method is supposed to learn overlapped two-level hierarchy among output classes. The first level is the \textbf{\textit{sparse mixture}} and second level contains several \textbf{\textit{sparse experts}}. A sparse expert is a cluster of classes that is a subset of the whole classes, and we allow each class to belong to more than one expert (non-exclusive). To generate the top-k classes, the sparse mixture enables a fast and dynamic selection of the right expert according to context vector $h$. And then the selected sparse expert allows a fast softmax computation over a small subset of the classes.

The framework is illustrated in Figure \ref{fig:overview_fig} and Algorithm 1 depicts our method, which contains two major components: (1) the sparse mixture/gating network indicates the sparse mixture and enables the selection of a top-1 expert, and (2) the sparse experts that are pruned from full softmax with group lasso. We also leverage a loading balance term to balance the utilization of different experts, and the mitosis training scheme to scale it to a larger number of experts. The final objective will be a combination of task-specific loss $\mathcal{L}_{task}$, group lasso loss $\mathcal{L}_{lasso}$ and some loading balance regularization losses $\mathcal{L}_{load}$ and $\mathcal{L}_{expert}$. The $\lambda$ indicates the corresponding weight. 

\begin{minipage}[t]{0.5\textwidth}
  \vspace{0pt} 
  \begin{algorithm}[H]
    \caption{DS-Softmax}
    \KwData{hidden representation $h$ and label $y$}
    \Parameter{gating $U$ and experts $W$}
    \While{training not converge}{
    $\mathcal{L}_{task} = \delta(O(h), y)$\;
    $\mathcal{L}_{all} = \mathcal{L}_{task}+\mathcal{L}_{lasso}+\mathcal{L}_{load} + \mathcal{L}_{expert} $\;
    $U = U - \alpha \frac{\partial}{\partial U} \mathcal{L}_{all}(x,y;W, U) $\;
    $W = W - \alpha \frac{\partial}{\partial W} \mathcal{L}_{all}(x,y;W, U) $\;

    \If{$\mathcal{L}_{task} \; < \; \text{threshold } t $}{
      \ForEach{$W^{(k)}_{c} \in W$}{
       \If{$\|W^{(k)}_{c}\|_2<\gamma$}{$W^{(k)}_{c} = 0$}
      }
    }
    }
    \label{alg:overview} 
  \end{algorithm}
\end{minipage}%
\hfill
\begin{minipage}[t]{0.45\textwidth}
  \vspace{0pt}
  \begin{algorithm}[H]
    \caption{Mitosis Training}
    \textbf{Initialize: } Gating $U^{(1)}$ and experts $W^{(1)}$ where there are only two experts\;
    $t = 1$ \;
    \While{$2^t < K$}{
        $U, W\gets$ run Algorithm 1 with $U^{(t)}$ and $W^{(t)}$ as initialization\;
        $U^{(t+1)} \gets$ randomly initialize $U^{(t+1)}$ with twice the number of experts\;
        $\tilde{W}\gets \text{add Gaussian noises to } W$\;
        $W^{(t+1)} \gets [\tilde{W}, \tilde{W}]$, i.e. double the experts by concatenation\;
        $t=t+1$ \;
    }
    \label{alg:mitosis}
  \end{algorithm}
\end{minipage}

\paragraph{Sparse mixture.} The first level of sparsification is a sparse gating, which is designed to find the right expert given the context vector $h$. To facilitate faster inference, only a \textit{single} most suitable expert is selected. To be more specific, suppose we have $K$ experts. Given the context vector $h$ and gating network weight $U$, the gating values $G_k(h)$, $k=1,...,K$, are calculated and normalized prior to the selection as shown in Eq.~\ref{eq:topone_gate}. And then we only maintain the largest gating value while set all other gates to be zero. More specifically, 
\begin{equation}
\label{eq:topone_gate}
G_{k}(h) =  
\begin{cases}
\frac{\exp (U_k h)}{\sum_{k'} \exp (U_{k'}  h)}, & \text{if } k = \argmax_i \frac{\exp (U_k h)}{\sum_{k'} \exp (U_{k'}  h)}, \\
0, & \text{otherwise}.
\end{cases} 
\end{equation}
Where $U \in \mathbb{R}^{K\times d}$ is the weighting matrix for group selection, and only the top-1 expert is selected. Eq.~\ref{eq:topone_gate} still allows the gradient to be back-propagated to whole $W^g$ due to normalization. It is worth noting that although our sparse gating network is similar to the one in \cite{shazeer2017outrageously}, when only top-1 expert is selected, our formulation has a valid gradient while their formulation does not.

Given the sparse gate, we can further compute the probability of class $c$ under the context $h$ as:
\begin{equation}
\label{eq:temperature}
O(h) = p(class=c|h) = \frac{\exp (\sum_k G_{k}(h) W^{(k)}_{c} h)}{\sum_{c'} \exp (\sum_k G_{k}(h) W^{(k)}_{c'} h)},
\end{equation}
where $W^{(k)} \in \mathbb{R}^{N \times d}$ is softmax embedding weight matrix for the $k$-th expert. Gating values can be interpreted as an inverse temperature term for final categorical distribution produced by the selected expert $k$~\citep{shazeer2017outrageously}. A smaller $G_k$ gives a more uniform distribution and a larger $G_k$ makes sharper one, and this can be adjusted automatically according to the context. It is worth noting that during the inference, we only need to compute single selected expert given the rest are zeros, and select top-k on the expert with a subset classes. The training is end-to-end, w.r.t. the task-specific loss function, i.e. $\mathcal{L}_{task} = D(O(h), y)$. In practice, we found pre-training all layers and just re-learn the softmax layer can achieve faster convergence. The task-specific end-to-end training allows our model to be consistent in both training and inference, as compared to the post-approximation methods.

\paragraph{Sparse expert.} The second level sparsification is making each expert sparse. We want each expert to contain only a small subset of whole classes, which means it should output a categorical distribution over vocabulary where most entries are zeros. To obtain a sparse expert, we start by initializing an expert as a full softmax that covers all classes and apply group lasso $\mathcal{L}_{lasso} = \lambda_{lasso} \; \sum_{k} \sum_{c} \| \hat{W}^{(k)}_{c} \|_2$ to iteratively prune out irrelevant classes. This regularization term actively prunes embedding vectors that in each expert once their $\ell_2$ norm is smaller than the pre-defined threshold $\gamma$. When heavily regularized, there will be many classes pruned out of each expert, leading to a set of sparse experts. Moreover, we include the expert level group lasso loss $\mathcal{L}_{expert} = \lambda_{expert} \; \sum_{k} \sqrt{\sum_c \| W^{(k)}_{c} \|_2^2}$ term so that each class is encouraged to exist in only one or a few experts. 

\paragraph{Loading balance.} To achieve a better speedup, balanced utilization of experts is necessary. We denote the number of final classes in expert $k$ as $|v_k|$, and the number of total classes as $|V|$. The utilization ratio $u_k$ indicates the probability of an expert being selected with a given dataset. For example, if model is run 10,000 times and $k$-th expert is selected for 100 times, then the utilization ratio $u_k$ is 0.01. The overall speedup is calculated as $|V| / (\sum_k (|v_k| * u_k) + k)$. Therefore, it is not desirable to have an unbalanced load since the model can degenerate to a single big softmax which leads to less speedup. To address this issue, we add the loading balance loss $\mathcal{L}_{load} = \lambda_{load}\; \text{CV} \left(\sum_{h \in H(x)} G_{k}(h) \right) ^ 2$ that encourages a more balanced utilization of experts similar to \citep{shazeer2017outrageously}. 

\begin{figure}[!t]
    \centering
    \includegraphics[width=\textwidth]{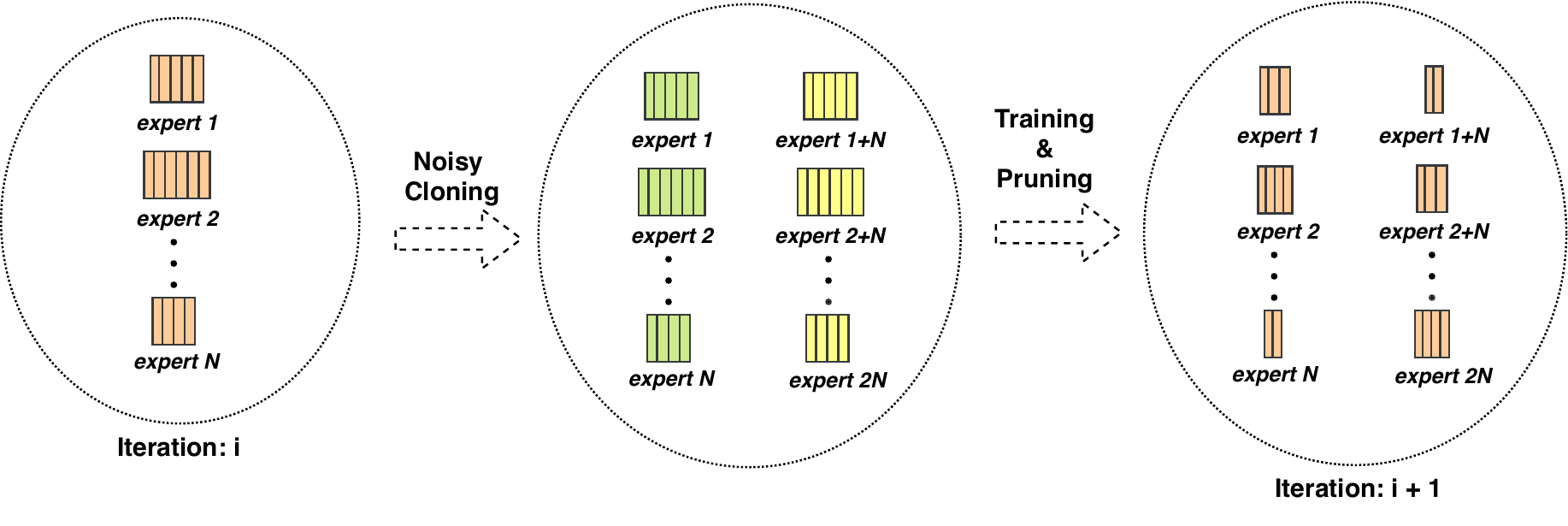}
    \caption{The mitosis training strategy: the sparsity is inherited when a parent expert produce offspring, reducing the memory requirements for training with large number of experts.}
    \label{fig:mitosis}
\end{figure}

\paragraph{Mitosis Training.} The mixture model requires initialization with full space, which introduces a memory bottleneck. This suggests that during training, the DS-Softmax requires $K$ times the memory as a regular softmax layer, which limits the number of experts used in of DS-Softmax. Therefore, we present an novel training scheme, mitosis training, to train mixture model with efficient memory usage. Mitosis training is a strategy to progressively increase the number of experts during the training. We start the training with a smaller number of experts. Once it converges, we split each expert into two identical ones and repeat the same training procedure with the initialized model. At the time of splitting/cloning one expert into two, the expert is already relatively sparse and smaller than the full softmax, it would require a much smaller memory consumption as the case without mitosis train. An illustration of the mitosis training can be found in Fig.~\ref{fig:mitosis} and algorithm can be found in Algorithm 2.

\paragraph{Complexity Analysis.} Here we analyze the inference computation complexity for the proposed softmax inference. The proposed method consists of two-step computation: (1) a sparse gating to choose an expert, which has $\mathcal{O}(K)$ complexity given $K$ experts; and (2) a small-scale softmax from the selected sparse expert to compute the sparse categorical distribution, which has an average of $\mathcal{O}(Nm/K+K)$ complexity given a balanced set of experts and a class/word on average belongs to $m$ experts, assuming the utilization of words is similar. The real speedup is calculated based on utilization and sparsity. 

\section{Experiments}
We present our empirical evaluations on both real and synthetic tasks in this section. Firstly, we create one synthetic task with two-level hierarchy and test our model's ability to learn the hierarchical structure. Secondly, we consider three real tasks, natural language modeling, neural machine translation, and Chinese handwritten character recognition. Both theoretical speedup (reduction in FLOPs) and real device latency (on CPU) are reported in Table ~\ref{tab:real}. Finally, some ablations and case study are present to better understand what the model has learned. For the baselines, we mainly compare to the conventional full softmax and recently proposed SVD-Softmax \citep{shim2017svd} and D-Softmax \citep{chen2015strategies, grave2016efficient}. Hyper-parameter details are shown in Appendix~\ref{app:hp_tuning}. 

\begin{figure*}[t]
  \centering
  \subfigure[Synthetic Data Generation]{\includegraphics[width=0.28\textwidth]{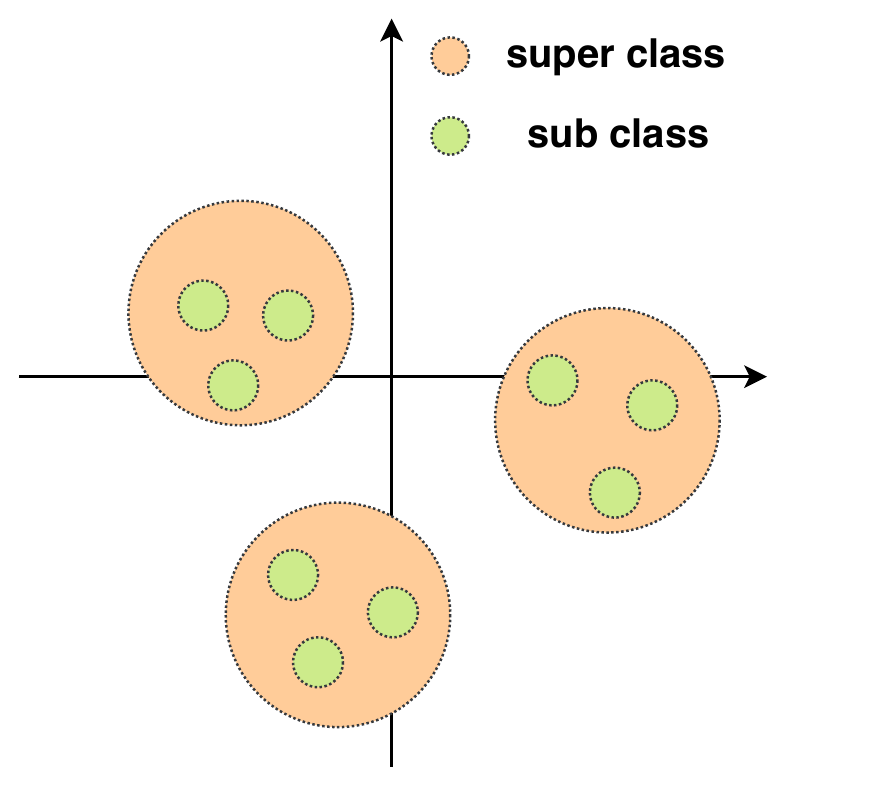}\label{fig:toy_data}}\qquad
  \subfigure[Results on 10 x 10]{\includegraphics[width=0.28\textwidth]{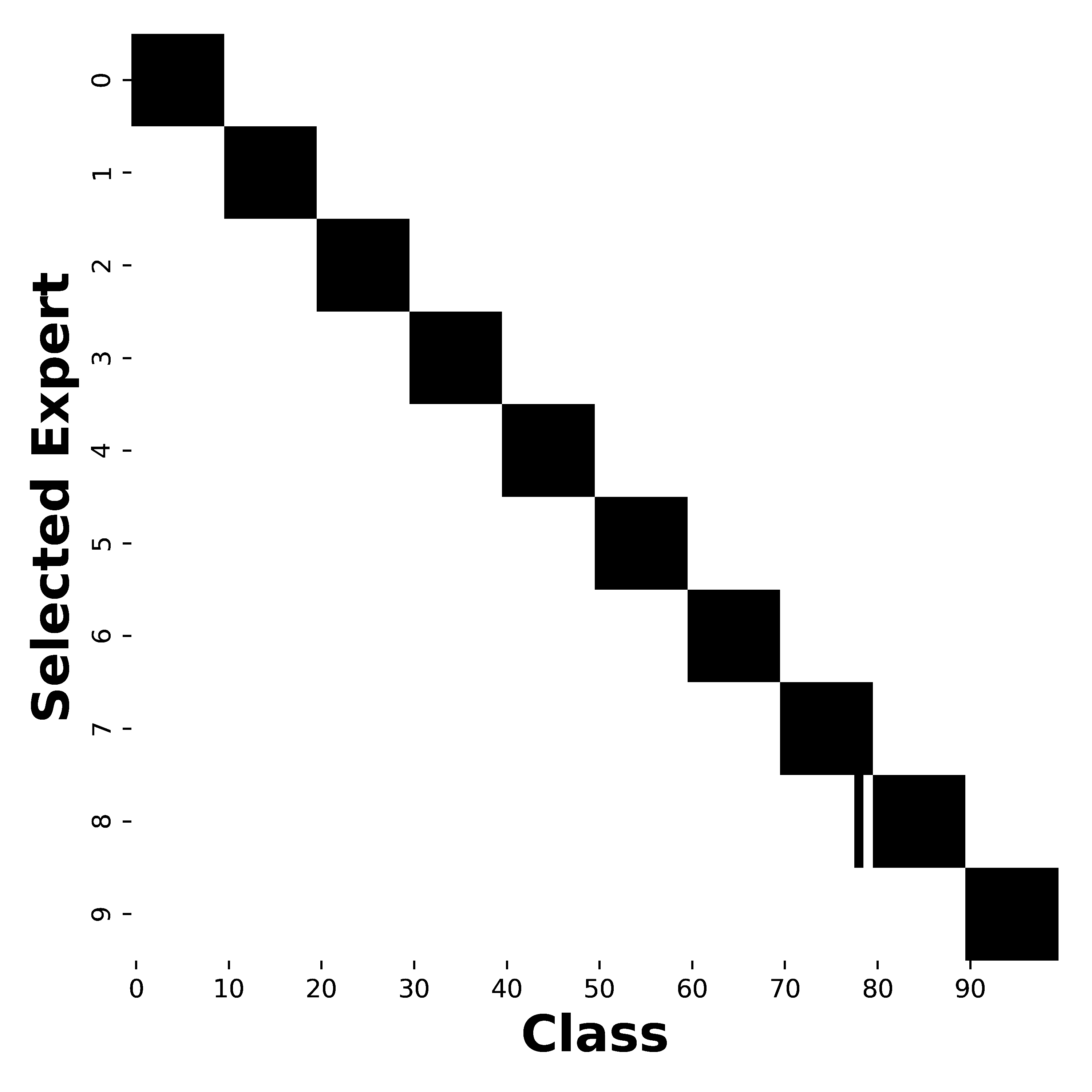}\label{fig:toy_100}}\qquad
  \subfigure[Results on 100 x 100]{\includegraphics[width=0.28\textwidth]{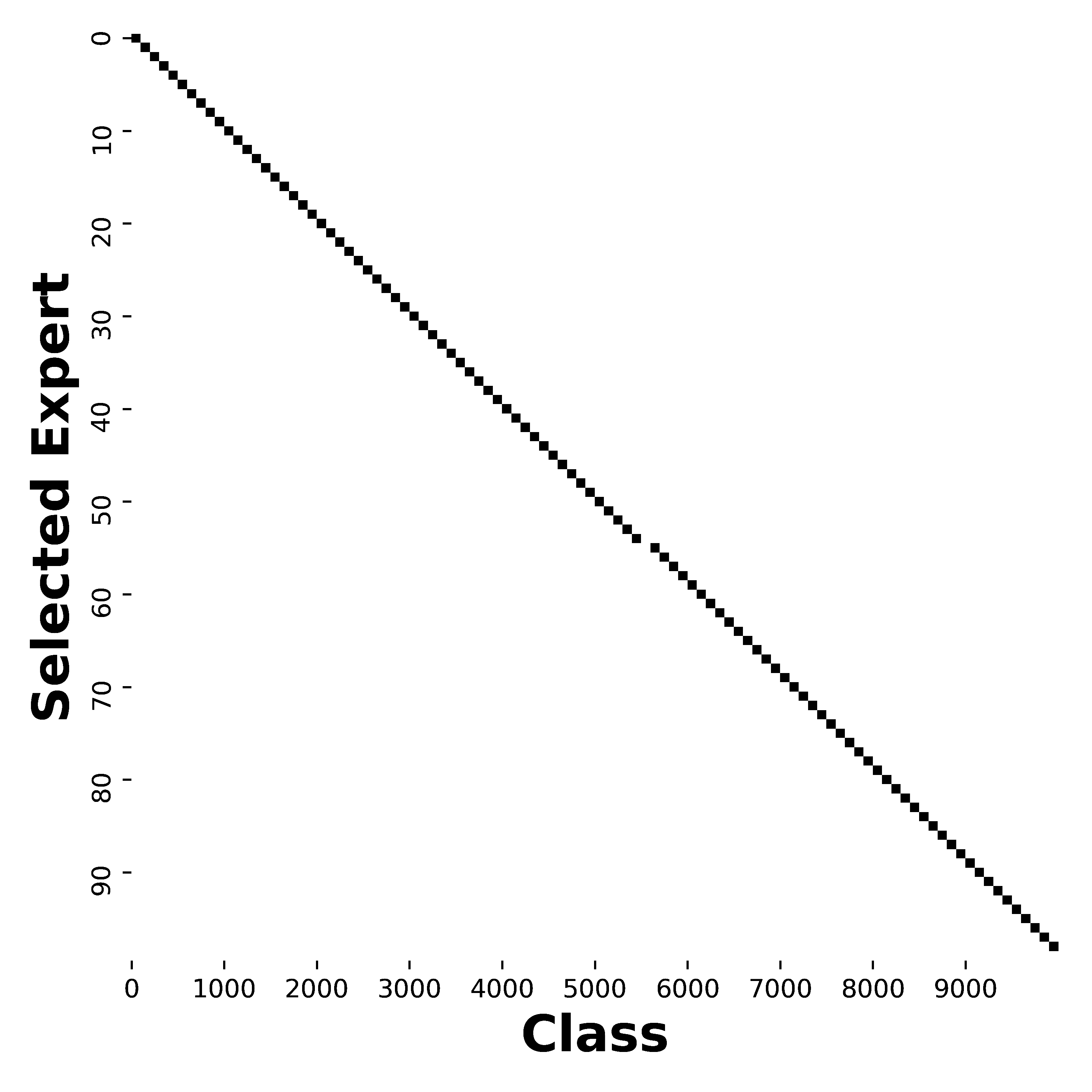}\label{fig:toy_10000}}
  \caption{(a) Illustration of synthetic data. The input is generated inside sub cluster (green circle) and its corresponding label is the sub cluster. The super cluster information is not present during training. (b) and (c) Results on tasks with 10x10 and 100x100 sizes. The x-axis indicates sub cluster and y-axis shows the expert. Black means this expert is handling this sub cluster. The order of x-axis is arranged through their super cluster information (e.g., in 10x10 size problem, first 10 sub classes are belonged to one super cluster, and so on).}\label{fig:animals}
\end{figure*}

\subsection{Synthetic Task}
A two-level hierarchy synthetic dataset is constructed to test our model. For sanity check and visualization purpose, we make sure the ground-truth hierarchy in the synthetic data without overlapping. As illustrated in Figure~\ref{fig:toy_data}, data points are organized with hierarchical centers, multiple sub classes belong to one super class. Detail of generation can be found in Appendix~\ref{app:synthetic}.

We treat the coordinates of a data point $x_j$ as features and the sub cluster membership of the data point as the target. We construct a two-layer Multi-layer Perception (MLP) with DS-softmax as the final layer for the task. We investigate the captured hierarchy by examining how sub clusters are distributed through experts. As mentioned, each expert only contains a subset of output classes, because class level pruning is conducted during training. We illustrate the remaining classes in each expert in Fig.~\ref{fig:toy_100} and Fig.~\ref{fig:toy_10000} for 10x10 and 100x100 sizes respectively. We find DS-Softmax can perfectly capture the hierarchy. We do further ablation analysis on the results on 10x10 synthetic as shown in Appendix~\ref{app:synthetic} to study the effect of each additional loss. As we can see, all the loss terms discussed above are important to our model.

\begin{table*}[t]
\small
\centering
\caption{Comparison with SVD-softmax and D-Softmax on real device latency. The ``ms'' indicates the latency in microseconds. ``FLOPs'' indicates FLOPs speedup. The value indicates the task performance. In PTB, Wiki-2 and CASIA, the value indicates the top-1 accuracy. In En-Ve, it means the BLEU score. D-Softmax cannot speedup in CASIA as the frequency of the classes is the same. }
\begin{tabular}{l|cc|ccc|ccc|ccc}
\toprule
\multicolumn{1}{c|}{Task} & \multicolumn{2}{c|}{Full} & \multicolumn{3}{c|}{SVD-10} & \multicolumn{3}{c|}{D-Softmax} & \multicolumn{3}{c}{DS-64 (Ours)} \\
 & Value & ms & Value & FLOPs & ms & Value & FLOPs & ms & Value & FLOPs & ms\\ \midrule
PTB & 0.252 & 0.73 & 0.251 & 5.00$\times$ & 0.18 & 0.245 & 2.00$\times$ & 0.36 & \textbf{0.258} & \textbf{15.99$\times$} & \textbf{0.05} \\ 
Wiki-2 & 0.257 & 3.07 & 0.255 & 5.38$\times$ & 0.60 & 0.256 & 2.00$\times$ & 1.59 & \textbf{0.259} & \textbf{23.86$\times$} & \textbf{0.15}\\
En-Ve & \textbf{25.2} & 1.91 & 25.1 & 5.06$\times$ & 0.42 & 24.8 & 2.00$\times$ & 0.98 & 25.0 & \textbf{15.08$\times$} & \textbf{0.13} \\
CASIA & \textbf{90.6} & 1.61 & 90.2 & 2.61$\times$ & 0.68 & - & - & - & 90.1 & \textbf{6.91$\times$} & \textbf{0.25} \\ 
\hline
\bottomrule
\end{tabular}
\label{tab:real}
\end{table*}


\subsection{Language Modeling}
Language modeling is a task whose goal is to predict the next word given the context. For a language such as English, a large vocabulary is present and softmax can be a bottleneck for inference efficiency. We use two standard datasets for word level language modelling: PennTree Bank (PTB) \citep{marcus1994penn} and WikiText-2 \citep{merity2016pointer}, where the output dimensions are 10,000 and 33,278 respectively. Standard two-layers LSTM model \citep{gers1999learning} with 200 hidden size is used\footnote{https://github.com/tensorflow/models/tree/master/tutorials/rnn/ptb}. We use top-K accuracy as our metric as it is a common metric \citep{chen1998evaluation} in natural language modeling especially in a real application when the extrinsic reward is given, such as voice recognition. Top 1, Top 5 and Top 10 accuracies on testing set are reported. Here, we compare both full softmax and FGD \citep{zhang2018navigating}. To enable a fair comparisons, we report the performances of FGD at different performance levels. We demonstrate that 15.99$\times$ and 23.86$\times$ times speedup (in terms of FLOPs) can be achieved with 64 experts without loss of accuracy, or even a slight improvement in some cases, as shown in Table ~\ref{tab:lm_result}. Moreover, without the constraint that each word has to exist in at least one expert, our model can achieve 34.78$\times$ and 90.84$\times$ speedup at similar performances. Qualitative result of PTB demonstrates the semantic meaningful clustering of classes are found, shown in Appendix~\ref{app:lm_result}. 

\subsection{Neural Machine Translation \& Chinese Character Recognition}
Neural machine translation task is also commonly used for softmax speedup evaluation. We use IWSLT English to Vietnamese dataset \citep{Luong-Manning:iwslt15} (the output vocabulary size is 7,709) and evaluate performance by BLEU score \citep{papineni2002bleu} with greedy searching. The BLEU is assessed on the testing set. A vanilla softmax model is seq2seq \citep{sutskever2014sequence} and implemented using TensorFlow\footnote{https://github.com/tensorflow/nmt} \citep{abadi2016tensorflow}. As shown in Table \ref{tab:real}, our model can achieve 15.08$\times$ speedup with similar BLEU score.

Beyond language applications, the efficiency of the proposed method is also demonstrated in classification tasks, such as Chinese handwriting character recognition task. We use the offline and special characters filtered CASIA dataset \citep{liu2011casia}. CAISA is a popular Chinese character recognition dataset with around four thousand characters. Unlike language related tasks, the distribution output class distribution is uniform/balanced. Two-thirds of the data is chosen for training and rest for testing. Our model can achieve significant (6.91$\times$) speedup on this task as shown in Table \ref{tab:real}.

The additional results on NMT and handwriting character recognition are available in Appendix~\ref{app:hp_tuning}.

\subsection{Real Device Comparison}
We further evaluation the efficiency of the proposed method on real device: a machine with two Intel(R) Xeon(R) CPU @ 2.20GHz, and 16G memory. All tested models are re-implemented using Numpy to ensure a fair comaprison. One configuration of SVD-Softmax~\cite{shim2017svd} is evaluated, i.e. SVD-10, which uses 10\% dimension for final evaluation in their preview window and window width is 16. Indexing and sorting are computationally heavy for SVD-softmax with Numpy implementation. One configuration of Differentiated(D)-Softmax is compared here, despite that their main focus is on training speedup \citep{chen2015strategies}. D-Softmax is selected instead of adaptive-softmax because they have same performance on CPU speedup \citep{grave2016efficient}. The words are sorted by their frequency, and the first quarter and second quarter utilize the same embedding size and half embedding size. The tail uses a quarter embedding size. For example, in PTB, we split the words into buckets (2500, 2500, 5000) and embedding sizes are (200, 100, 50). For a fair comparison, we report latency without sorting and indexing for SVD-softmax. However, regards to full softmax, D-Softmax, DS-Softmax, full latency is reported. The latency results are shown in Table~\ref{tab:real}, and we observe that DS-Softmax can achieve significantly better theoretic speedup ($4.5\times$ better on Wiki-2) as well as lower latency ($2.9\times$ faster on Wiki-2). Moreover, compared to D-Softmax, we find our learned hierarchy can achieve much better speedup without loss of performance. 

\subsection{Mitosis Training}
Here we demonstrate the efficiency of mitosis training on PTB language modeling task. The model is initialized with 2 experts, and clones to 4, 8, 16, 32 and 64 experts sequentially. Cloning happens for every 15 epochs and pruning starts 10 epochs after cloning. In the end, the model only requires at most 3.25$\times$ memory to train DS-64 model and achieve similar performance, significantly smaller than original 64-fold memory. The performance illustration is shown in Appendix~\ref{app:mitosis}.

\begin{table*}[t]
\small
\begin{minipage}{0.56\textwidth}
\caption{Further comparisons of word level natural language modelling on PTB and WikiText-2, which have 10,000 and 33,278 words respectively. "L" and "H" mean the low and high precision. The 'Speed' indicates the reduction of FLOPs. '$*$' means pruning without at least one copy constraint.}
\begin{tabular}{l|c|ccc|c}
\toprule
\multirow{2}{*}{Task} & \multirow{2}{*}{Method} & \multicolumn{3}{c|}{Testing Accuracy} & \multirow{2}{*}{Speed} \\ 
& & Top 1 & Top 5 & Top 10 &  \\ \midrule
\multirow{5}{*}{\thead{PTB \\ (10,000)}} & Full & 0.252 & 0.436 & 0.515 & - \\
& FGD-H & 0.249 & 0.430 & 0.502 & 1.31$\times$ \\
& FGD-L & 0.227 & 0.391 & 0.455 & 6.76$\times$ \\
& DS-8 & 0.257 & 0.448 & \textbf{0.530} & 2.84$\times$ \\
& DS-64 & \textbf{0.259} & \textbf{0.450} & 0.529 & 15.99$\times$ \\
& $\text{DS-64}^*$ & 0.258 & 0.449 & 0.528 & \textbf{34.78$\times$} \\\midrule

\multirow{5}{*}{\thead{WIKI-2 \\ (33,278)}} & Full & 0.257 & 0.456 & 0.533 & - \\
& FGD-H & 0.254 & 0.437 & 0.509 & 4.79$\times$ \\
& FGD-L & 0.233 & 0.380 & 0.442 & 22.45$\times$ \\
& DS-8 & 0.259 & \textbf{0.459} & \textbf{0.536} & 3.52$\times$ \\
& DS-64  & 0.259 & 0.458 & 0.533 & 23.86$\times$ \\
& $\text{DS-64}^*$ & \textbf{0.260} & 0.458 & 0.534 & \textbf{90.84$\times$} \\\hline \bottomrule
\end{tabular}
\label{tab:lm_result}
\end{minipage}
\hspace{0.025\textwidth}
\begin{minipage}{0.42\textwidth}
\caption{Evaluation of applying post-approximation methods on the learned experts from DS-Softmax, which can further speedup the inference.}
\begin{tabular}{l|c|c|c}
\toprule
Task & Method & Top 1 & Speed \\ \midrule
\multirow{6}{*}{\thead{WIKI-2 \\ (33,278)}} & Full & 0.257 & - \\
& DS-2 & 0.258 & 1.83$\times$ \\
& SVD-10 & 0.255 & 5.38$\times$ \\
& D2+S10 & 0.255 & 9.64$\times$ \\
& DS-64 & \textbf{0.259} & 23.86$\times$ \\
& SVD-50 & 0.256 & 1.72$\times$ \\
& D64+S50 & 0.255 & \textbf{32.77$\times$} \\\hline\bottomrule
\end{tabular} 
\label{table:orth}
\end{minipage}
\end{table*}

\subsection{Post-approximation of Learned Experts}
To speedup softmax inference, most existing methods are based on post-approximation of a learned and fixed softmax \citep{shim2017svd, chen2018learning, mussmann2017fast}. In DS-Softmax, we can consider each expert as an individual softmax with a subset of whole classes. This suggests that the post-approximation technique \citep{shim2017svd} can be applied upon DS-Softmax. To demonstrate this, two experiments are conducted. One is applying SVD-10 to DS-2. Another is applying SVD-50 (top 50\% in the preview window) to DS-64, where SVD is applied upon to expert with more than one thousand classes. The higher percent in SVD is used for DS-64 because there are fewer remaining classes in each expert. Table~\ref{table:orth} shows combination achieves better performance.

\section{Conclusion}

In this paper, we present {\it doubly sparse: a sparse mixture of sparse experts} for efficient softmax inference. Our method is learning-based and adapts softmax for fast inference. It learns a two-level overlapping class hierarchy. Each expert is learned to be only responsible for a small subset of the output class space. During inference, our method first identifies the responsible expert and then performs a small-scale softmax computation by the expert. Our experiments on several real-world tasks have demonstrated the efficacy of the proposed method.

\newpage
\bibliography{ref}
\bibliographystyle{unsrt}
\appendix
\newpage
\counterwithin{figure}{section}
\counterwithin{table}{section}

\section{Synthetic Data Experiment}
\label{app:synthetic}

\paragraph{Data Generation: } The data is generated as following procedures. First, one centroid for super class $c^{super}$ is generated from one Gaussian with mean as zeros and variances as $d^3 I$. Then the sub class is generated with corresponding super class but smaller variances. Finally, the data points are generated around the sub class. The details are: 

\begin{align}
\label{eq:toy_data}
c_{i}^{\text{super}} & \sim \mathcal{N}(0, d^3 I), \\
c_{j}^{\text{sub}} & \sim \mathcal{N}(c_{i}^{\text{super}}, d^2 I), \\
x_{j}^{\text{input}} & \sim \mathcal{N}(c_{j}^{\text{sub}}, d I).
\end{align}

\paragraph{Ablation Result: } Each mentioned is removed for each experiment to test its importance in our module. The result is demonstrated in following Fig~\ref{fig:analysis}. 

\begin{figure*}[h]
  \centering
  \subfigure[No Group Lasso]{\includegraphics[width=0.25\textwidth]{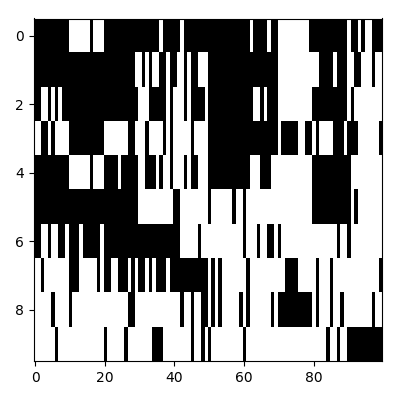}}\qquad
  \subfigure[No Expert Group Lasso]{\includegraphics[width=0.25\textwidth]{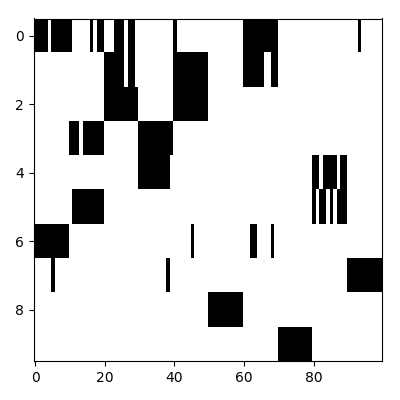}}\qquad
  \subfigure[No Balancing]{\includegraphics[width=0.25\textwidth]{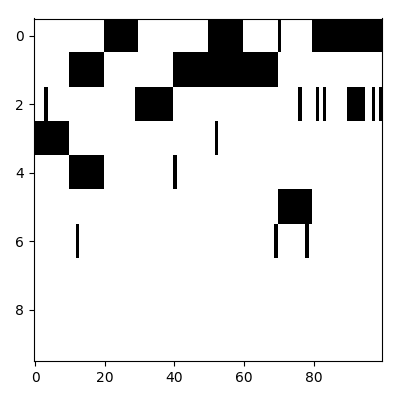}}
  \caption{Ablation analysis of each loss component by removing it. (a), (b) and (c) illustrate the model trained without group lasso, expert level group lasso and balancing factor, respectively. The original result is Fig.~\ref{fig:toy_100}, and they share the same axis .}\label{fig:analysis}
\end{figure*}

\newpage
\section{Real Tasks Experiment Details}
\label{app:hp_tuning}

In terms of experiment setup, we leave the task-specific matters for later, here we present details on our model setup. The proposed DS-Softmax layer can be trained jointly with other layers in an end-to-end fashion. For real tasks, we find it is easier to first pre-train the whole model with conventional softmax, and replace the softmax layer with DS-Softmax and retrain the new layer while keeping others fixed, with Adam \citep{kingma2014adam}. For hyper-parameters, $\lambda_{load}$ and threshold $\gamma$ in pruning are fixed for all tasks as 10 and 0.01 respectively. $\lambda_{lasso}$ and $\lambda_{expert}$ share the same value and are tuned using the following strategy: starting with zero and increasing exponentially until it decreases the performance in validation. The reported performance is on an independent testing dataset.

This part also presents the experimental details for IWSLT machine translation (Table C.1) and Chinese character recognition dataset (CASIA) (Table C.2). 

\begin{table}[h]
\small
\centering
\caption{Neural machine translation results on IWSLT English to Vietnamese and the vocabulary size is 7,709.}
\begin{tabular}{l|c|c|c}
\toprule
Task & Method & Bleu Score & Speedup \\\midrule
\multirow{5}{*}{\thead{IWSLT \\ En-Ve \\ (7,709)}} & Full & 25.2 & - \\
& DS-8 & 25.3 & 4.38x \\
& DS-16 & 25.1 & 6.08x \\
& DS-32 & \textbf{25.4} & 10.69x \\
& DS-64 & 25.0 & \textbf{15.08x} \\\hline\bottomrule
\end{tabular}
\label{table:nmt_result}
\end{table}

\begin{table}[h]
\small
\centering
\caption{Image classification results on CASIA. There are 3,740 different characters inside dataset.}
\begin{tabular}{l|c|c|c}
\toprule
Task & Method & Accuracy & Speedup \\\midrule
\multirow{5}{*}{\thead{CASIA \\ (3,740)}} & Full & 90.6 & - \\
& DS-8 & \textbf{90.8} & 1.77x \\
& DS-16 & 90.2 & 2.82x \\
& DS-32 & 89.9 & 4.72x \\
& DS-64 & 90.1 & \textbf{6.91x} \\\hline\bottomrule
\end{tabular} 
\label{table:casia}
\end{table}

\newpage
\section{Qualitative Result in Language Modeling}
\label{app:lm_result}

We demonstrate the redundancy and word frequency pattern in Figure D.1, where the redundancy indicates the number of experts contains such word. We find words with higher frequency will appear in more experts. This is a similar phenomenon as the topic models in \citet{blei2003latent, wallach2006topic}, and similar fact that more frequent words require higher capacity model \citep{chen2015strategies}. We manually interrogate the smallest expert in such a model, where 64 words remain\footnote{The words existing in more than experts are filtered.}. The words left in such expert is semantically related. Three major groups are identified, which are \textit{money}, \textit{time} and \textit{comparison}, shown in following: 

\begin{itemize}[noitemsep,topsep=0pt,parsep=0pt,partopsep=0pt]
\item \textbf{Money}: million, billion, trillion, earnings, share, rate, stake, bond, cents, bid, cash, fine, payable.
\item \textbf{Time}: years, while, since, before, early, late, yesterday, annual, currently, monthly, annually, Monday, Tuesday, Wednesday, Thursday, Friday.
\item \textbf{Comparison}: up, down, under, above, below, next, though, against, during, within, including, range, higher, lower, drop, rise, growth, increase, less, compared, unchanged.
\end{itemize}
\vspace{10pt}
\begin{figure}[h]
  \centering
  \includegraphics[width=0.35\textwidth]{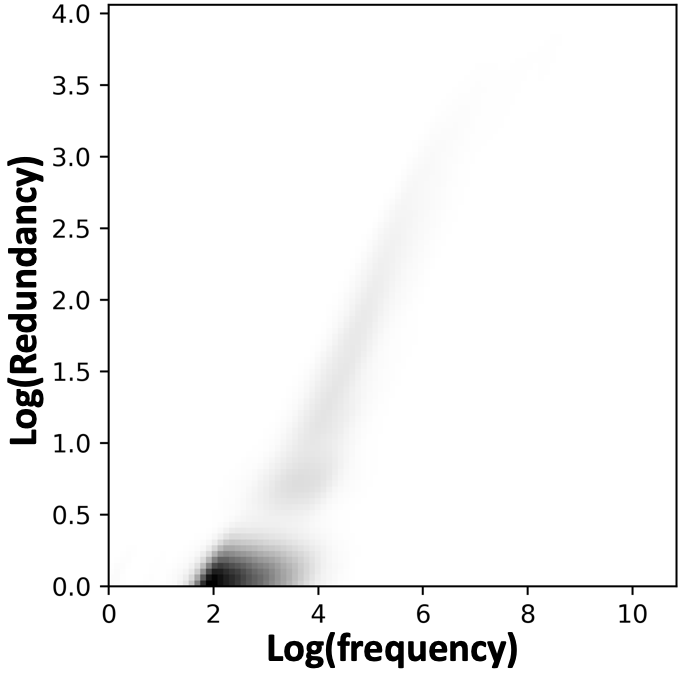}
  \caption{Uncertainty and Redundancy: A heatmap to demonstrate the correlation between word frequency and its redundancy. The x-axis is the log of word frequency and the y-axis is the log of number of expert containing this word (called Redundancy). Darker color indicates higher density. }\label{fig:analysis2}
\end{figure}

\newpage
\section{Mitosis Training Result}
\label{app:mitosis}

This section we present the mitosis training result in natural language modelling task with PTB dataset. For each 15 epoch of training, the model clones the number of experts into double, e.g. from 2 to 4, and 4 to 8. We demonstrated how the memory usage is changing during the training in the following figure. The performance at each cloning point is reported as well. 

\begin{table}[h]
\small
\centering
\caption{Language Modelling Result During Mitosis Training}
\begin{tabular}{l|c|ccc|c}
\toprule
\multirow{2}{*}{Task} & \multirow{2}{*}{Method} & \multicolumn{3}{c|}{Testing Accuracy} & \multirow{2}{*}{Speed} \\ 
& & Top 1 & Top 5 & Top 10 &  \\ \midrule
\multirow{5}{*}{\thead{PTB \\ (10,000)}} & DS-2 & 0.253 & 0.439 & 0.521 & 1.78$\times$ \\
& DS-4 & 0.255 & 0.442 & 0.519 & 2.04$\times$ \\
& DS-8 & 0.253 & 0.445 & 0.525 & 2.44$\times$ \\
& DS-16 & 0.257 & 0.449 & 0.530 & 4.99$\times$ \\
& DS-32 & 0.256 & 0.447 & 0.530 & 9.10 $\times$ \\
& DS-64 & 0.258 & 0.448 & 0.528 & 14.47 $\times$ \\\hline\bottomrule
\end{tabular} 
\label{table:casia}
\end{table}

\begin{figure}[h]
  \centering
  \includegraphics[width=0.35\textwidth]{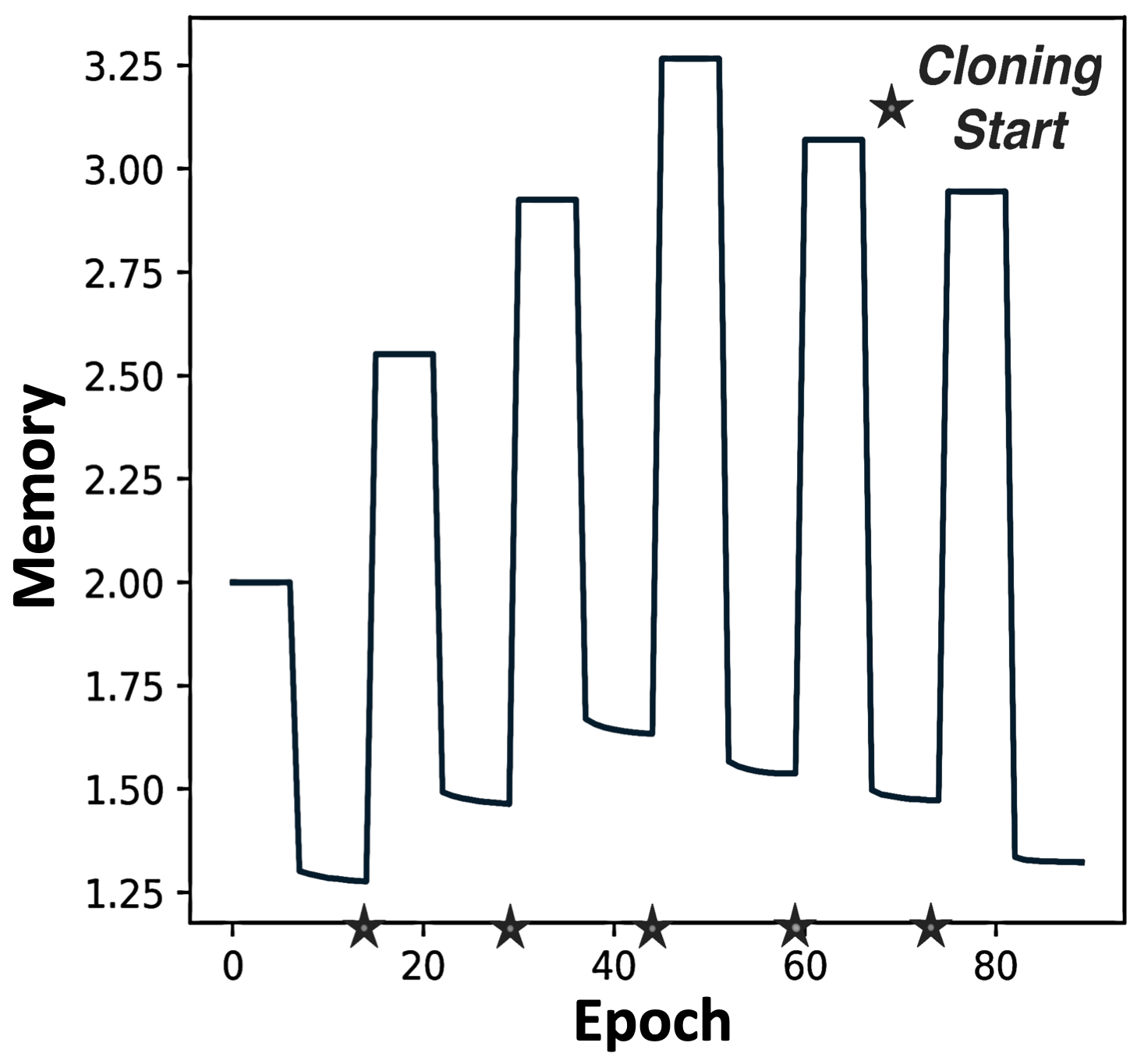}
  \caption{Illustration of required memory to train DS-64 starting with DS-2. The y-axis is the memory comparing to one full softmax. The Cloning Start icon means where the cloning happens. }\label{fig:analysis2}
\end{figure}

\end{document}